\documentclass[11pt,a4paper]{article}
\usepackage{preamble}

\begin{document}
	\maketitle
	\begin{abstract}
	   Language models are typically evaluated on their success at predicting the distribution of specific words in specific contexts. Yet linguistic knowledge also encodes relationships between contexts, allowing inferences between word distributions. We investigate the degree to which pre-trained Transformer-based large language models (LLMs) represent such relationships, focusing on the domain of argument structure. We find that LLMs perform well in generalizing the distribution of a novel noun argument between related contexts that were seen during pre-training  (e.g., the active object and passive subject of the verb \emph{spray}), succeeding by making use of the semantically-organized structure of the embedding space for word embeddings. However, LLMs fail at  generalizations between related contexts that have not been observed during pre-training, but which instantiate more abstract, but well-attested structural generalizations (e.g., between the active object and passive subject of an arbitrary verb). Instead, in this case, LLMs show a bias to generalize based on linear order. This finding points to a limitation with current models and points to a reason for which their training is data-intensive.\footnote{The code and datasets used for the experiments reported here are available at \url{https://github.com/clay-lab/structural-alternations}.}
	\end{abstract}
	
	\section{Introduction}
        Competent speakers of a language know how likely a word $w$ is to appear in a specific context $C$, which we define as the sequence of words that surrounds a certain position in a sentence. Any speaker of English knows, for example, that the word \emph{kids} is reasonably likely to occur in the context \textit{the \underline{\hspace{2em}} hit the ball} and that \emph{ball} is likely to appear in the context \textit{the kid hit the \underline{\hspace{2em}}}. We call this kind of knowledge about specific words in specific contexts ``Type 0 knowledge.''
        \begin{quote}
            \textbf{Type 0 knowledge}: For a specific word $w$, $w$ occurs with probability $p$ in context $C$.
        \end{quote}
        By itself, such Type 0 knowledge is woefully incomplete. Competent speakers need to be able to generalize their knowledge of distributions across certain pairs of contexts:
        \begin{quote}
            \textbf{Type 1 knowledge}: For all words $w$, if $w$ occurs in $C$, $w$ can also occur in $C'$.
        \end{quote} 
        Type 1 knowledge tells a speaker that appearance in one context is predictive of appearance in another. Not only do English speakers know that \emph{kids} or \emph{ball} are likely to occur in the specific contexts given above, they also know that occurrence in these contexts is predictive of occurrence in others, where the determiner is changed, other nouns are used, adverbs are added, etc., essentially any other context in which  the relation of the predicted noun to the verb \textit{hit} remains. The relevant bit of Type 1 knowledge thus implicates sensitivity to linguistic abstractions like grammatical role: a noun that is a likely subject of a particular sentence with active transitive \textit{hit} will also be a likely subject of another sentence in which it is also the subject of \textit{hit}.  In fact, Type 1 knowledge goes further still: any noun that is a likely object of an active transitive \emph{hit} is a likely subject of passive  \emph{hit}. One kind of Type 1 generalization can thus be characterized in terms of thematic role: both the object of active transitive \emph{hit} and the subject of passive \emph{hit} are mapped to the theme role of an hitting event, and this thematic role imposes certain selectional preferences on the choice of noun. Similarly, both the subject of active transitive \emph{hit} and the object of the \emph{by}-phrase in passive \emph{hit} are mapped to the agent role, which gives rise to other selectional preferences that apply to both contexts. 
    	\begin{exe}
    		\ex{\begin{xlist}
    			\ex{{[}Kids{]}\subs{\textsc{agent}} hit {[}balls{]}\subs{\textsc{theme}}.}
    			\ex{{[}Balls{]}\subs{\textsc{theme}} are hit by {[}kids{]}\subs{\textsc{agent}}.}
    		\end{xlist}}
    	\end{exe}
        In this case, the relevant bit of Type 1 knowledge says that contexts for a word that put it in the same thematic role with respect to \textit{hit} support Type 1 generalization.
        
        In fact, a competent speaker has knowledge of even more abstract generalizations over contexts: 
        \begin{quote} 
            \textbf{Type 2 knowledge}: For all contexts $C$ and $C'$, if  $C$ and $C'$ are appropriately related, then for all $w$, if  $w$ can occur in $C$, $w$ can also occur in $C'$. 
        \end{quote} 
        Characterizing Type 2 knowledge requires defining what it means for two contexts to be appropriately related, and for the purposes of this paper we will focus on contexts in  which the same thematic role is assigned. The Type 1 knowledge just discussed concerning the arguments of \textit{hit} is a specific instance of such a relation. The relevant Type 2 knowledge is a meta-level generalization, holding across argument positions in active and passive sentences regardless of the specific verb or the roles it assigns. For example, object experiencer verbs like \emph{bother} show a thematic mapping distinct from the one found with \emph{hit}: active transitive subjects are the theme, while the object is the experiencer (and hence necessarily animate). This means that likely subjects and objects of active experiencer verbs differ substantially from those of verbs like \emph{hit}. Nonetheless, when each type of verbs is passivized, the same Type 2 generalization holds: the role assigned to the passive subject, and hence which nouns are likely subjects, is the same as the role of the corresponding active object.
        
        The range of pairs of contexts that share thematic roles, and hence fall under the Type 2 generalization under discussion, goes well beyond voice alternations like active and passive. For instance, in a cleft, any argument of a verb may appear in a position to the left of its unmarked (active) use.
    	\begin{exe}
    		\ex{\begin{xlist}
    			\ex{It is {[}kids{]}\subs{\textsc{agent}} who hit {[}balls{]}\subs{\textsc{theme}}.}
    			\ex{It is {[}balls{]}\subs{\textsc{theme}} that {[}kids{]}\subs{\textsc{agent}} hit.}
    		\end{xlist}}
    	\end{exe}
        And once again, English speakers understand that the argument out of its canonical position bears the same thematic relation to the event, and shows the same preferences for possible nouns. In other words, speakers of a language understand not only relationships between specific structures with specific lexical items, but also higher level productive relationships between structures.
        
        Recent work has demonstrated that Large Language Models (LLMs) exhibit sensitivity to a wide range of subtle regularities of linguistic form (\citealt{LinzenDupouxGoldberg16,MarvinLinzen18,Goldberg19,Wilcoxetal2018}, \emph{inter alia}). Yet because of the vast quantities of data on which LLMs are trained, it is difficult to determine the degree to which these results derive from Type 0, Type 1, or Type 2 knowledge. Given sufficient quantities of training data, a LLM can behave as though it abides by Type 1 or Type 2 knowledge, so long as it has received explicit and consistent evidence about each context. But in order to be robust in the face of infrequent words and rare constructions, true Type 1 and Type 2 knowledge will be crucial. In this paper, we assess the presence of Type 1 and Type 2 generalizations in LLMs, specifically in the domain of the mapping between structural position and thematic role as diagnosed by selectional preferences.
        
        Focusing on the BERT family of transformer\hyp based language models (BERT, \citealt{Devlinetal:2019}; DistilBERT, \citealt{Sahnetal:2019}; and RoBERTa, \citealt{Liuetal:2019}), our strategy is to introduce novel tokens into the models' vocabularies. We fine\hyp tune pre\hyp trained LLMs in a single structural context that links each novel token to a specific participant role \citep[cf.][]{KimSmolensky:2021,PettyWilsonFrank:2022a}, and then examine predictions in unobserved but related structural contexts for the novel tokens. We consider a different set of verbs and a wider range of structural contexts than were investigated in this previous work, and investigate what underlies the models' behavior in more detail by examining the embeddings of the novel words. In addition, while this work has focused on Type 1 knowledge, we investigate Type 2 knowledge as well, in experiment 2. Since the LLMs have no experience with the novel tokens we teach them in the test contexts, whatever success they have must be due to knowledge acquired during pre\hyp training. This allows us to address our general question about LLMs' knowledge of relationships between contexts by focusing on specific questions about their knowledge of argument structure.
        
        We find that LLMs demonstrate relatively robust Type 1 knowledge: they generalize the use of novel argument tokens across different structures for existing verbs, suggesting that their knowledge of position\hyp role mappings is abstract enough to recognize a notion of construction over which generalizations can be stated. However, deeper probing reveals that they do not exhibit human\hyp like Type 2 knowledge of generalization across construction types. We show this by introducing a new verb into the models' vocabulary that has novel selectional preferences. Though LLMs learned these novel preferences and generalized them to structures that are superficially similar to those in the fine-tuning data, they failed to consistently generalize them to other contexts in a structurally\hyp defined, human\hyp like manner. Instead, their generalizations were derived through heuristics based on linear order, yielding high performance in structures where the relative linear order of arguments and main verb is preserved (e.g., polar questions), and poor performance in structures where the relative linear order of arguments is reversed (e.g., canonical passives). We conclude that the models represent abstract structures, but that the Type 2 generalizations they seem to represent rely on surface properties of the training data.
    
    \section{Experiment 1: Assessing Type 1 generalization with novel nouns}
        Experiment 1 examines the degree to which LLMs show evidence of Type 1 knowledge by examining position\hyp role mappings\slash selectional preferences across constructions. We explore whether fine\hyp tuning on a novel noun, presented in a single context, will allow the LLM to predict its appearance in a related context. Our primary type of relation among contexts derives from argument structure alternations, where the same set of thematic roles can be realized with distinct structures. We adopt this strategy following \citet{PettyWilsonFrank:2022a}, who examine Type 1 knowledge with novel arguments of dative verbs. We focus instead on the \emph{spray}\slash\emph{load} alternation. In (\ref{spray-load}) we see that \emph{spray}\slash\emph{load} alternating verbs can occur in a theme-object (TO) structure, with the theme as the object and the goal as prepositional argument; or in a goal-object (GO) structure, with the goal as the object and the theme as the prepositional argument.
        \begin{exe}
           \ex{\emph{spray}\slash\emph{load} alternation:\begin{xlist}
                \ex{I sprayed the \textsc{thax} onto the door.\\
                    I sprayed the paint onto the \textsc{gorx}.\\\push(theme-object (TO) structure)}
                \ex{I sprayed the door with the \textsc{thax}.\\
                    I sprayed the \textsc{gorx} with the paint.\\\push(goal-object (GO) structure)}
            \end{xlist}}
            \label{spray-load}
        \end{exe}
        We introduced a novel token that can appear in each of the non-subject argument positions: \emph{\textsc{thax}} for theme arguments and \emph{\textsc{gorx}} for goal arguments.\footnote{For both experiments 1 and 2, we added the novel tokens to the LLMs' vocabularies such that they were not split into subword tokens during tokenization (verified during each run). Our scripts also verified during each run that all training and test sentences were tokenized identically when using a tokenizer with the added tokens and the same tokenizer without the added tokens (except in the position of the added tokens).} We fine-tuned LLMs on a small set of simple sentences containing \emph{spray}\slash\emph{load} verbs in one of their alternating forms: TO or GO. All of the sentences for a given run were headed by one of two \emph{spray}\slash\emph{load} verbs: \emph{spray} or \emph{load}.
        
        This meant that there were four distinct fine-tuning sets, one for each combination of the syntactic structure (TO or GO) and lexical verb. Every example in these sets contained exactly one novel token (as in (\ref{spray-load})), with the remaining arguments drawn from a small set of semantically plausible nouns. This yielded a total of 12 fine-tuning examples per set.\footnote{An anonymous reviewer notes that this makes our fine-tuning sets quite small, which may affect their ability to learn a good representation of the novel tokens. However, we note that \citet{KimSmolensky:2021} find that LLMs of the sort we investigate are able to glean relevant distributional information from a fine-tuning set consisting of as little as two short sentences. In addition, the models achieved high levels of success in experiment 1. For experiment 2, where less success is seen, our fine-tuning sets were larger (288 sentences). Thus, we do not believe the small size of the fine-tuning sets in experiment 1 is likely to have had a negative impact on our results.}
        
        Our validation set includes four subparts, whose performance was averaged: the fine-tuning dataset evaluated without dropout, sentences instantiating the opposite form of the alternation (e.g., \emph{spray} GO used \emph{spray} TO in the validation set, and vice versa), the dataset with the other fine-tuning verb for that alternation (e.g., \emph{spray} GO used \emph{load} TO in the validation set, and vice versa), and the dataset in passive (e.g., active \emph{spray} GO used passive \emph{spray} GO in the validation set, and vice versa). The validation set was thus constructed to penalize overfitting to our fine\hyp tuning sentences at the expense of sentences with various other structures compatible with our verbs.
        
        To assess an LLM's sensitivity to Type 1 generalizations, our test set considered the fine-tuned model's predictions on sentences that differed from the fine-tuning data along three structural dimensions: (i) the argument structure alternation, (ii) passivization, and (iii) other syntactic transformations, including clefting, polar and (matrix\slash embedded) wh-question formation, negation, raising, relative clause formation, and particle shift. In contrast to \citet{PettyWilsonFrank:2022a}'s test data, which included only four test structures (crossing the alternation and passivization), the way we combined our three dimensions gave rise a much broader range of 78 test structures. We show a subset of these here, with the positions corresponding to each novel word filled in with the expected token.
        \begin{exe}
           \ex{
                \begin{xlist}
                    \ex{The man sprayed the \textsc{thax} onto the \textsc{gorx}. \push(TO active)}
                    \ex{The \textsc{gorx} was sprayed with the \textsc{thax}. \push(GO passive)}
                    \ex{It was the man that sprayed the \textsc{thax} onto the \textsc{gorx}. \push(cleft subj TO active)}
                    \ex{It was the \textsc{thax} that the \textsc{gorx} was sprayed with. \\\push(cleft P-obj GO passive)}
                    \ex{Which \textsc{thax} was the \textsc{gorx} sprayed with? \push(matrix \emph{wh}-P-obj GO passive)}
                    \ex{I wonder which \textsc{gorx} the man seems to have sprayed onto the \textsc{thax}.\\\push(emb \emph{wh}-obj, emb raising TO active)}
                \end{xlist}
            }
            \label{short-examples}
        \end{exe}
        To examine generalization to sentences with verbs distinct from those used in the fine\hyp tuning data, but which share their thematic roles and participate in the alternation, we used the following verbs in test data, grouped into relevant lexical semantic subclasses \citep[based on][]{Pinker:1989}.
        \begin{exe}
            \ex{Test verbs:\begin{xlist}
                \ex{\emph{spray}-type: \emph{spray}, \emph{shower}, \emph{dab}, \emph{rub}}
                \ex{\emph{load}-type: \emph{load}, \emph{stock}, \emph{pack}, \emph{stuff}}
            \end{xlist}}\label{verbs}
        \end{exe}
        Though the broad thematic roles of the verbs within each class are shared, they differ to some degree at a fine-grained level that will impact their selectional preferences.\footnote{The \emph{spray}-type subclass of \emph{spray}\slash\emph{load} verbs typically take themes that refer to substances (rather than discrete entities) that are distributed over a surface, whether as a particulate (\emph{spray}, \emph{shower}) or as a kind of paste (\emph{dab}, \emph{rub}). The \emph{load}-type subclass typically take themes that refer to discrete countable entities and involve placing those upon or inside the goal, with \emph{pack} and \emph{stuff} further specifying that the goal is being filled against the limits of its capacity.}
        This led to an evaluation set of 5,616 sentences (78 structural contexts instantiated with different non-target nouns and the lexical verbs from (\ref{verbs})).
        
        During evaluation, novel nouns were replaced by a \textsc{mask} token, and the predicted probabilities for these positions were extracted from the LLM. A prediction for a position was taken to be accurate if the model assigns higher probability to the expected novel token than the unexpected one.\footnote{We also repeated experiment 1 using verbs that participate in the \emph{dative} alternation and testing on our 78 structures (though due to the lack of a preposition in the double object dative structure, some of our test structures were ambiguous with dative verbs). Results were comparable to those we report for \emph{spray}\slash\emph{load} verbs.}
        
        \subsection{Setup and hyperparameters}
            We used the Hugging Face \texttt{transformers} library \citep{huggingface} to fine-tune the BERT base uncased, DistilBERT base uncased, and RoBERTa base models. Prior to fine-tuning, we froze all model parameters except the embeddings of the novel tokens \emph{\textsc{thax}} and \emph{\textsc{gorx}}, which were randomly initialized for each run. We use a masked language modeling objective, evaluated at the positions associated with the masked novel tokens. The learning rate for all runs was $0.001$. Batches contained the full set of 12 (\emph{spray} or \emph{load}) sentences. We train until convergence on validation set loss, using early stopping with a patience of 30. Results are averages across five different random initializations of the novel token embeddings.
        
        \subsection{Results}
            All models were highly successful in learning the distribution of the novel tokens in the fine-tuning datasets: average accuracy was $99.6\%$ for DistilBERT, $100.0\%$ for BERT and $98.1\%$ for RoBERTa. 
            On the test set, the models were again fairly successful: 
            $82.6\%$ for DistilBERT, $89.2\%$ for BERT and $81.0\%$ for RoBERTa.\footnote{The accuracies we report for test data are conditionalized; we only consider those positions in which the prediction is accurate for the related fine-tuning structure. Thus, for an LLM that had been fine-tuned on theme object sentences, only if the fine-tuned model is accurate in the \textsc{thax} position in (\ref{short-examples}a) would success in the \textsc{thax} positions in the other examples in (\ref{short-examples}) be included in the accuracy computation, and similarly for the \textsc{gorx} positions. Our accuracy metric thus reports the proportion of sentences in a particular condition for which the LLM predicts the correct placement of the novel token(s), provided it has learned how to place them in the fine-tuning structure. Conditionalizing the accuracy allows us to interpret it as a measure of generalization. While the high level of training set accuracy makes this choice of little importance for experiment 1 (and does not affect BERT at all), we use this measure to remain consistent with experiment 2, where it is of more consequence.}
            The high generalization accuracy points to the ability of LLMs to do Type 1 generalizations: the distribution of a novel noun in one context can be extended to a distinct context. Not all generalizations were equally successful, however.
            \begin{table}
                \centering\tiny
                \begin{tabular}{cccc}
                    \toprule
                                                &                        & TO        & GO\\
                    \midrule
                    \multirow{2}{*}{RoBERTa}    & goal (\textsc{gorx})   & \tp{80.6} & \tp{89.2}  \\
                                                & theme  (\textsc{thax}) & \tp{79.8} & \tp{74.2}  \\
                    \midrule
                    \multirow{2}{*}{BERT}       & goal  (\textsc{gorx})  & \tp{85.7} & \tp{96.7}  \\
                                                & theme  (\textsc{thax}) & \tp{92.9} & \tp{81.3}  \\
                    \midrule
                    \multirow{2}{*}{DistilBERT} & goal  (\textsc{gorx})  & \tp{82.1} & \tp{96.6}  \\
                                                & theme  (\textsc{thax}) & \tp{90.1} & \tp{61.8}  \\
                    \bottomrule
                \end{tabular}
                \caption{\label{tab:thematicacc} Mean conditional accuracy by model, novel token thematic role, and fine-tuning structure. Chance performance is 50\%.}\vspace{-4mm}
            \end{table}
            
            Consider first generalization performance by thematic role, shown in Table \ref{tab:thematicacc}. Here, we find that RoBERTa shows an advantage for goal arguments across all fine-tuning regimens.
            In contrast, BERT and DistilBERT show a bias towards adjacent objects: they predict objects that were adjacent to the verb in the fine-tuning data more accurately (i.e., \emph{\textsc{thax}} in TO and \emph{\textsc{gorx}}
            in GO).
            
            Next, we break down performance by movement type, reported in Table \ref{tab:movementacc}.\footnote{An anonymous reviewer asks if there were other interesting patterns that depend on the structure of the test sentences. For reasons of space, we are unable to go into details for each of the 78 structure types in our test set. Our full set of results is publicly available at our GitHub repo (\url{https://github.com/clay-lab/structural-alternations}). Here, we present a breakdown that best illustrates the most salient properties of our test set results.} Long-distance dependencies in transformational generative grammar are modeled as a constituent ``moving'' from one place in a syntactic structure to another. Different subtypes of movement can be distinguished: A-movement refers to displacement of a constituent into a canonical argument position, such as passive and raising, which target the subject position. \abar{A}-movement (``A-bar movement'') involves displacement of a constituent into non-argument positions, including clefting, \emph{wh}-movement, relativization, and so on. The worst performance occurs in sentences with both movement types (e.g., \emph{which \textsc{gorx} was the \textsc{thax} sprayed onto?}, which has passive A-movement, and \emph{wh} \abar{A}-movement). This effect appears to be superadditive.
            
            \begin{table}[t]
                \centering\tiny
                \begin{tabular}{cccc}
                    \toprule
                                               & \multirow{2}{*}{A-movement?} & \multicolumn{2}{c}{\abar{A}-movement?}\\
                                               &                              & \cmark    & \xmark\\
                   \midrule
                   \multirow{2}{*}{RoBERTa}    & \cmark                       & \tp{77.7} & \tp{85.7}\\
                                               & \xmark                       & \tp{82.1} & \tp{86.3}\\
                   \midrule
                   \multirow{2}{*}{BERT}       & \cmark                       & \tp{87.0} & \tp{93.9}\\
                                               & \xmark                       & \tp{88.8} & \tp{93.2}\\
                   \midrule
                   \multirow{2}{*}{DistilBERT} & \cmark                       & \tp{79.3} & \tp{89.3}\\
                                               & \xmark                       & \tp{82.1} & \tp{90.0}\\
                   \bottomrule
                \end{tabular}
                \caption{\label{tab:movementacc} Mean conditional accuracy by model, A-movement, and \abar{A}-movement.}\vspace{-2mm}
            \end{table}
            
            A concern facing the LLMs we investigate is that their pre-training data does not give them direct access to structural generalizations beyond those based on linear properties of strings. On the other hand, human language use is characterized by structure-dependence. While \citet{lin-etal-2019-open}, \citet{hewitt-2019}, and \citet{jawahar-etal-2019-bert} present evidence for the encoding of hierarchical structure in transformer-based language models, that work does not directly address the question of generalization we explore here. To get at that question, we break down accuracy by the relative linear order of arguments compared to the fine-tuning data in Table \ref{tab:linearorderaccexp1}. Despite overall high performance, all models perform somewhat worse when the linear order of the novel tokens differs from the order in which they occurred in the fine-tuning sentences, as in the alternative argument structure or certain questions or clefts. Recall, though, that no fine-tuning sentence contains both novel tokens, so even generalizations based on the relative linear order of the arguments must be based on knowledge acquired during pre-training.
            
            \begin{table}[t]
                \centering\tiny
                \begin{tabular}{crr|r}
                    \toprule
                                & \multicolumn{1}{c}{same} & \multicolumn{1}{c}{reverse} & \multicolumn{1}{|c}{penalty} \\
                    \midrule
                    RoBERTa     & \tp{83.2}                & \tp{78.7}                   & \tablenum{4.5}\\
                    BERT        & \tp{92.5}                & \tp{85.8}                   & \tablenum{6.7}\\
                    DistilBERT  & \tp{87.5}                & \tp{77.8}                   & \tablenum{9.7}\\
                    \bottomrule
                \end{tabular}
                \caption{\label{tab:linearorderaccexp1} Mean conditional accuracy by linear order of arguments relative to the fine-tuning data.}\vspace{-4mm}
            \end{table}
        
        \subsection{Word embedding analysis}
            Our results show that LLMs are able to generalize the use of the novel arguments across a broad range of lexical and syntactic contexts to an overall high degree of accuracy, confirming the generalizability of \citet{PettyWilsonFrank:2022a}'s findings to a broader set of alternations, structures, and verbs. We now turn to the question of how the LLMs achieve this generalization by investigating learned embeddings of the novel tokens, the only part of the model updated during fine-tuning.
            
\begin{figure}[t]
                \centering
                \includegraphics[width=0.75\linewidth]{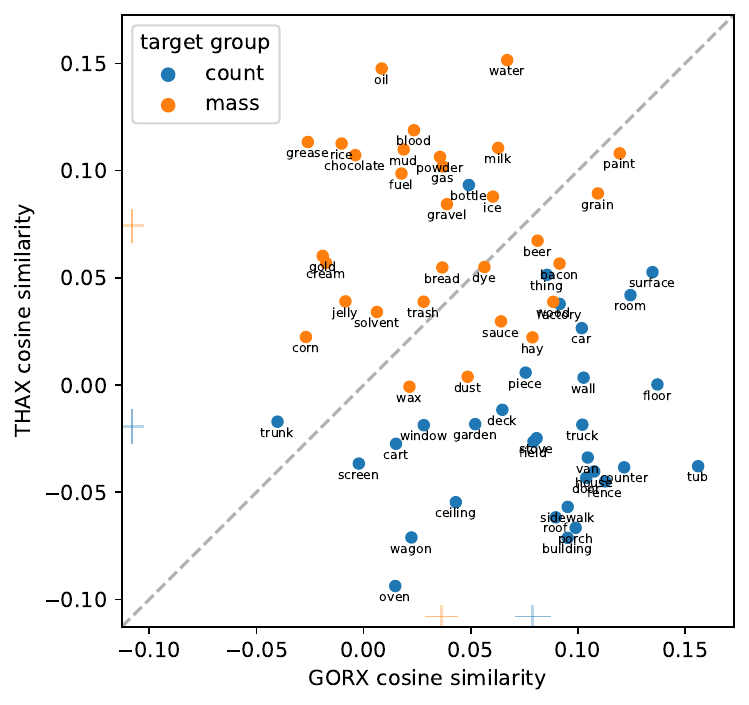}
                \caption{\label{fig:cossim1} Cosine similarities of count and mass nouns to novel tokens in a BERT model fine-tuned on TO sentences with \emph{spray}. The diagonal line indicates equal similarity to both novel tokens; higher values on an axis reflect greater similarity to that token. Ticks on axis margins indicate group means and standard errors.}\vspace{-4mm}
            \end{figure}

\begin{figure}[t]
                \centering
                \includegraphics[width=0.75\linewidth]{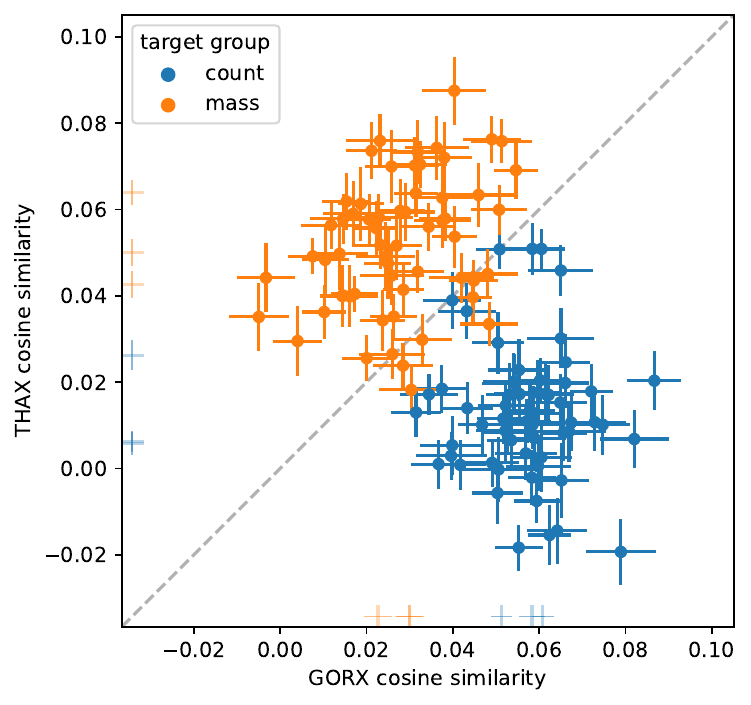}
                \caption{\label{fig:cossimsmulti} Mean cosine similarity of target groups of words to the novel tokens across 60 models fine-tuned with different verbs (\emph{spray}, \emph{load}) and structures (TO, GO). Each point represents the mean for a single model across words in the target group (count or mass).
                Bars represent standard errors.}\vspace{-4mm}
            \end{figure}

            Figures \ref{fig:cossim1} and \ref{fig:cossimsmulti} show plots from one representative model and across multiple models, respectively, of the cosine similarities (based on the pre-contextual word embedding layer) of the novel tokens to selected plausible themes and goals. Following \citet{cai2021isotropy} and \citet{timkey-van-schijndel-2021-bark}, we use all-but-the-top correction, removing the top 3 PCA dimensions, to account for anisotropy in the embedding space.\footnote{We thank an anonymous reviewer for alerting us to the  problem of anisotropy in the embedding space of transformer models, and the solution.} The learned embedding of \textsc{thax} is more similar to mass noun tokens that are plausible themes than to count noun tokens that are plausible goals, suggesting that the models have used information from the verb and syntactic context to induce a sensible meaning for the novel token. We see a similar pattern for goals: \textsc{gorx} is more similar to count nouns.
            
            As an anonymous reviewer points out, we might have expected no such consistency given our use of two distinct verbs (\emph{spray} and \emph{load}) for fine-tuning. The kinds of nouns that are appropriate themes and goals of \emph{spray} are different from those that are appropriate themes and goals of \emph{load}. Despite this, it appears that the models have instead learned rather general properties for the novel tokens (mass vs. count) instead of meanings that are idiosyncratic to particular verbs.
        
        \subsection{Investigating overgeneralization}
            The overall high accuracy for experiment 1 indicates that the models did not overgeneralize to sentences with two positions for the novel argument nouns. However, there are other ways in which overgeneralization might occur that our test dataset did not check. For instance, consider the following sentences:
            \begin{exe}
                \ex{\begin{xlist}
                    \ex{They loaded the hay with a pitchfork.}
                    \ex{They sprayed the hay with bugs.\\\push(e.g., using pesticide)}\label{revwith}
                \end{xlist}}
            \end{exe}
            In these cases, the meaning of the \emph{with}-phrase noun discourages interpreting it as the theme (though it is not impossible if a sufficiently fanciful scenario is envisaged). And in (\ref{revwith}), the proper passivization of this interpretation is even distinct from the goal-object passive.
            \begin{exe}
                \ex{\begin{xlist}
                    \ex{The hay with bugs was sprayed. $\neq$}
                    \ex{The hay was sprayed with bugs.}
                \end{xlist}}
            \end{exe}
            Given that our fine-tuning data is ambiguous between these different parses, as required by the use of novel tokens whose meanings cannot distinguish different senses like meanings of existing words that the models have seen in different contexts, an important question is how the models predictions fare in contexts that disambiguate in favor of one parse or another.
            
\begin{figure}[t]
                \centering
                \includegraphics[width=0.75\linewidth]{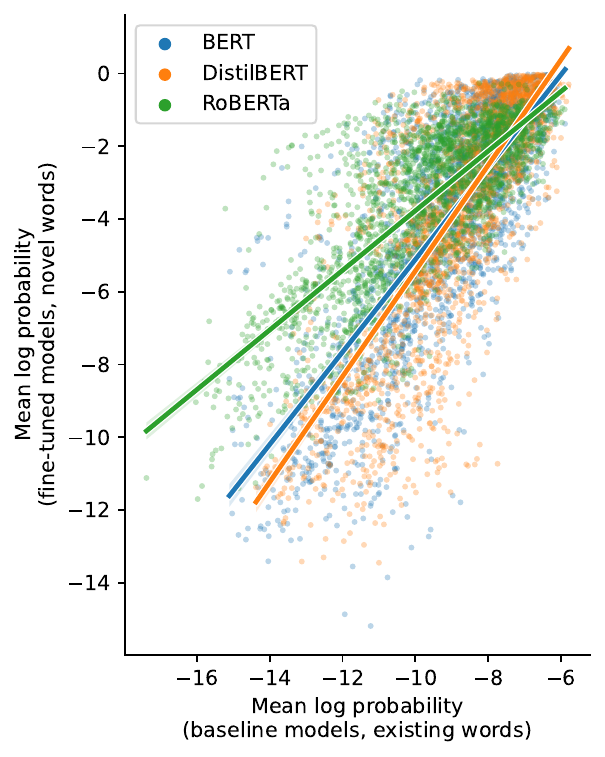}
                \caption{\label{fig:exp1overgen} Log probabilities of mass and count noun tokens for baseline models ($x$-axis) vs.\ for \textsc{thax} and \textsc{gorx} for fine-tuned models ($y$-axis). Each point represents a single mask position in a particular sentential context.}\vspace{-4mm}
            \end{figure}

            To examine this, we examined the models' predictions for \textsc{thax} and \textsc{gorx} in a variety of sentences with two PPs, like those in (\ref{exp1overgen}), using all of the $8$ verbs we evaluated our test datasets on.
            \begin{exe}
                \ex{\begin{xlist}
                    \ex{The [MASK] with the [MASK] was sprayed with the [MASK].}\label{twowith}
                    \ex{The [MASK] on the [MASK] was sprayed on the [MASK].}
                \end{xlist}}\label{exp1overgen}
            \end{exe}
            In (\ref{twowith}), the first \emph{with}-phrase must be interpreted as an NP modifier, while the second \emph{with}-phrase is ambiguous between an instrument and a theme interpretation. Given that even simple sentences like \emph{I sprayed the wall with paint} are technically ambiguous between an NP modifier parse and a theme parse, we compared predictions for \textsc{thax} and \textsc{gorx} in each mask position to the predictions of the non-fine-tuned versions of the models for the mass and count nouns we used in the discussion of cosine similarities in each position. Results are shown in fig. \ref{fig:exp1overgen}.
            
            We find that baseline predictions for tokens that the fine-tuned models consider similar to \textsc{thax} and \textsc{gorx} are highly correlated with their predictions across a variety of these partially disambiguating contexts (BERT: $r = 0.77$, $p < 2.2 \times 10^{-16}$; DistilBERT: $r = 0.79$, $p < 2.2 \times 10^{-16}$; RoBERTa: $r = 0.78$, $p < 2.2 \times 10^{-16}$).
            
            Thus, the models' predictions for words that serve as good themes or goals of \emph{spray}\slash\emph{load} verbs pre-fine-tuning are highly similar to their predictions for \textsc{thax} and \textsc{gorx} post-fine-tuning. In other words, the models do not seem to overgeneralize \textsc{thax} and \textsc{gorx}, but instead treat them like the pre-existing words they come to consider them similar to.
    
    \section{Experiment 2: Assessing Type 2 generalization with novel verbs}
        The results of experiment 1 show that LLMs can generalize the use of novel tokens to related structural contexts, demonstrating knowledge of position-role mappings that cuts across many constructions. Our exploration of the learned embeddings of the novel tokens suggests that the LLMs' ability to generalize in this way may be linked to semantic properties encoded in individual token embeddings. If the model knows that words represented in a certain embedding subspace are compatible with multiple structural contexts and it learns that a novel word's representation falls in that subspace, it will generalize the distribution of a word learned in one context to another. However, such a strategy does not suffice for Type 2 generalization, which requires the recognition of systematic relationships across contexts, independent of the specific properties of the contexts.  A model could succeed in experiment 1 by learning preferences for each context separately, so long as these are characterized by the same embedding subspace. We would then expect the syntactic complexity effect shown in Table \ref{tab:movementacc}, as cases with more types of movement will be less frequent in the training data and have less robust representations of selectional preferences. 
        
        Experiment 2 examines Type 2 generalization more directly. We modify the experiment 1 paradigm to introduce a novel verb token with novel selectional preferences. This renders the novel verb unlike existing ones, so the LLM cannot succeed by assimilating the novel verb to a known class of verbs with associated embedding subspace, as it did with experiment 1's novel nouns. Instead, success requires recognizing regularities that cut across all verbs, which will apply to the novel verb.
        
        We are not aware of studies that explicitly demonstrate a human capacity for this kind of generalization, though the following thought experiment suggests it would be possible. Suppose we invent a new video game which includes an action called \emph{blorking} that can be done only to an idiosyncratic set of objects---say mushrooms, clocks, and starfish. If we expose new gamers to sentences with \textsc{blork} only in the active voice as in (\ref{active-blork}), it seems clear that they will generalize their knowledge of its selectional preferences in the way that a Type 2 generalization would suggest, for example preferring the appropriate nouns in the passive subject position, as in (\ref{passive-blork}).
        \begin{exe}
            \ex{\begin{xlist}
                \ex{I just \textsc{blorked} the mushroom!}
                \label{active-blork}
                \ex{The mushroom was just \textsc{blorked}!}
                \label{passive-blork}
            \end{xlist}}
        \end{exe}
        Are LLMs capable of a similar feat?
        
        Note that we are not saying that people do not represent Type 0 and Type 1 generalizations: they clearly do \citep[e.g.,][]{Gleitman:1990,Gleitmanetal:2005}. However, we believe people represent Type 2 generalizations, too. Our question is simply whether models like BERT have such generalizations, or whether they are limited to the Type 0 and Type 1 generalizations we examined in Experiment 1.
        
        Experiment 2 focuses specifically on the generalization from active to passive sentences discussed above, where active transitive objects are subject to the same preferences as passive subjects, regardless of the specific choice of verb. Our fine-tuning dataset consists of 8 sentences with a novel verb used in the perfect, like the following.
        \begin{exe}
            \ex{The {[}subj{]} has always {\sc blorked} the {[}obj{]}.}
        \end{exe}
        Sentences differ only in the choice of preverbal adverb and presence (or not) of a sentence-final modifier. For each of \textsc{blork}'s argument positions, [subj] and [obj], we select 6 nouns in the LLM vocabulary as possible arguments. To ensure that assimilation to the selectional properties of an existing verb was not a viable strategy, we chose nouns that did not exhibit a clear preference for one position over the other prior to fine-tuning.\footnote{We did this as follows: for each model we randomly initialized the novel verb embedding. Across the set of 8 fine-tuning sentences as well as 5 more complex sentences, we compared the model's predictions in subject and object positions for all nouns from the SUBTLEX corpus \citep{BrysbaertNew:2009} of $>3$ characters that were tokenized as a single token in all models. From this set of predictions, we identified the 12 nouns with the lowest average sum of squared log odds ratios, i.e., $\sum_s \left(\log \frac{p(\text{sub}=n|s)}{p(\text{obj}=n|s)}\right)^2$. From these 12, we used nouns 0, 3, 4, 7, 10, and 11 as subject arguments, and the others as object arguments.}
        
        During fine-tuning, sentences were presented with both the novel verb and its argument nouns masked, as this maximized performance. As validation sets, we used the fine-tuning set with dropout disabled, as well as perfect passive, past-tense active, and past-tense passive sentences with the novel verb. We use these validation set subparts to ensure that the model did not overfit to active structures (via the inclusion of passive sentences), nor to perfect structures (via the inclusion of past tense sentences). We found fine-tuning until convergence did not always produce the best generalization, as overall validation loss would decrease despite overfitting to one of our validation set subparts. Instead, we fine-tuned for a maximum of 260 epochs ($=$ weight updates), with a minimum of 100 epochs and a patience of 30, a regimen that yielded the most success for BERT.\footnote{An anonymous reviewer notes that such a lengthy period of fine-tuning may be excessive, given the risk of overfitting, despite our attempts to address this risk via the modified loss term and the construction of our validation set. However, \citet{Mosbachetal:2020} evaluate BERT models' stability over fine-tuning, and ultimately recommend a ``hard-to-beat baseline'' for fine-tuning BERT that relies on many iterations and training until loss is (almost) zero. In addition, while we fine-tuned for a \emph{maximum} of 260 epochs, we always selected the model state which performed best on the validation set for evaluation, even if this was a state prior to 260 epochs (or prior to 100 epochs). Figure \ref{fig:exp2metrics} (discussed in more detail later) shows mean validation accuracy from each validation set subpart (perfect transitive (no dropout), perfect passive, past transitive, past passive) for each weight update\slash epoch over the course of fine-tuning until convergence. It is clear that the model does not overfit to active sentences at the expense of passive sentences; rather, its performance on passive sentences rises slowly but fails to achieve a high level of success.}
        
        Our test set consisted of a similar set of sentence types to those from experiment 1, which included passive sentences containing the verb \textsc{blork}. To form such sentences, we exploited the widely attested homophony of English past and passive participles in order to create passive sentences using the same, fine-tuned, novel verb token for \textsc{blorked}. 
        
        \subsection{Setup and hyperparameters}
            As in Experiment 1, we carried out fine-tuning on BERT, RoBERTa, and DistilBERT models. In order to investigate to what extent performance was reflective of properties of the LLM architecture under study rather than randomness in pre\hyp training, we additionally fine\hyp tuned 5 MultiBERT models. These are models whose architecture and pre\hyp training regimen are like BERT, but with different random initializations \citep{Sellametal:2021}.
            
            For fine-tuning, we initially attempted freezing all parameters except the novel token embedding as in experiment 1. However, this prevented the models from learning the unusual selectional preferences of the novel verb even for the fine-tuning data. This meant that a larger set of parameters needed to be modified during fine-tuning. However, unfreezing model parameters opens up the possibility of catastrophic forgetting \citep{mccloskey1989catastrophic, french1999catastrophic}, where a model's representations of existing tokens (i.e., the argument nouns) change to accommodate the fine-tuning data and lose their predictive utility outside of this data. For instance, if during fine-tuning a model assimilated its representations of the 6 selected subject arguments to animate nouns (even if they are inanimate), and its representations of the 6 selected object arguments to liquids (even if they are not liquids), it could succeed on the task by assimilating the novel verb to the existing verb \emph{drink}. However, this would reflect only a Type 1 generalization that is specific to \emph{drink} (and words like it). Such a strategy would have a negative effect outside of the fine-tuning dataset, as the selected subject and object nouns would now be predicted in inappropriate contexts.
            
            We explored a number of strategies to counter this kind of overfitting, while still allowing the novel selectional preferences to be learned. The most successful one involved unfreezing all model parameters while introducing a modification to the loss function, following \citet{Hawkinsetal:2019}. To cross-entropy, we add a penalty for Kullback-Leibler divergence between predictions of the fine-tuned model $M$ and those of the baseline model $B$ (i.e., post-pre-training but pre-fine-tuning). We extract a different random sample $S$ of 100 sentences from English Wikipedia and the BooksCorpus at each update, for which we calculate the summed KL divergence across all token positions $t$ in sentence $s \in S$.\footnote{Our random sample approximately matched the proportion of sentences drawn from each corpus to BERT's pre-training regimen, with 68\% drawn from English Wikipedia and 32\% drawn from the BooksCorpus. We used the \texttt{20200501.en} Wikipedia and BooksCorpus datasets from Hugging Face's \texttt{datasets} library \citep{datasets}.}
            \[
                L = L_{CE} + \lambda{\sum_{s\in S}}{\sum_{t \in s}} D_{KL}(p_M(t)||p_B(t))
            \]
            We used $\lambda=2.5$ as a scaling factor. Smaller values did not adequately address overfitting; larger ones prevented learning of the novel selectional preferences during fine-tuning. Experiment 2 used a learning rate of $0.0001$ to address instability found when unfreezing model parameters with the learning rate of $0.001$ from experiment 1.
            
\begin{figure}[t]
                \centering
                \includegraphics[width=\linewidth]{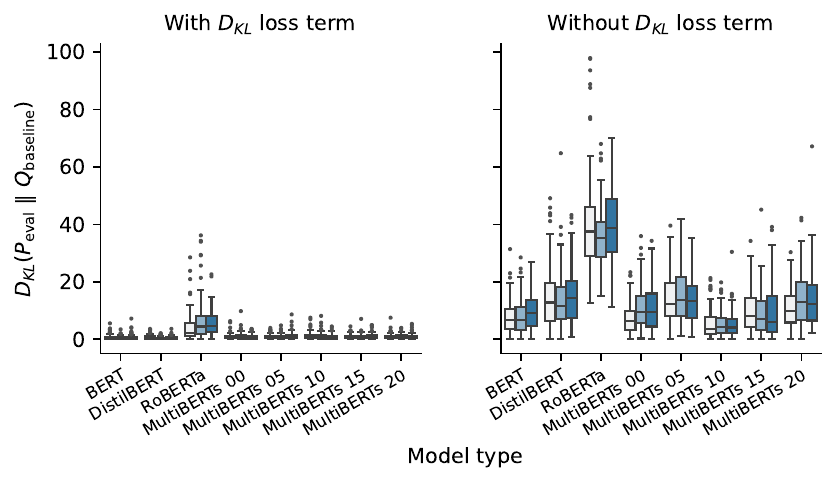}
                \caption{\label{fig:exp2-kldivs} $D_{KL}$ distributions for each fine-tuned model type compared to its pre-trained version in experiment 2, calculated on the basis of 100 examples randomly chosen from 10,000 selected from English Wikipedia and the BookCorpus. The left half of the plot shows $D_{KL}$ distributions for each model with our modified loss term, and the right half shows the $D_{KL}$ distributions for each model when using only the default cross entropy loss ($L_{CE}$). Each model has three associated distributions, one for each of the random seeds used in experiment 2.}\vspace{-4mm}
            \end{figure}

\begin{figure}[t]
                \centering
                \includegraphics[width=\linewidth]{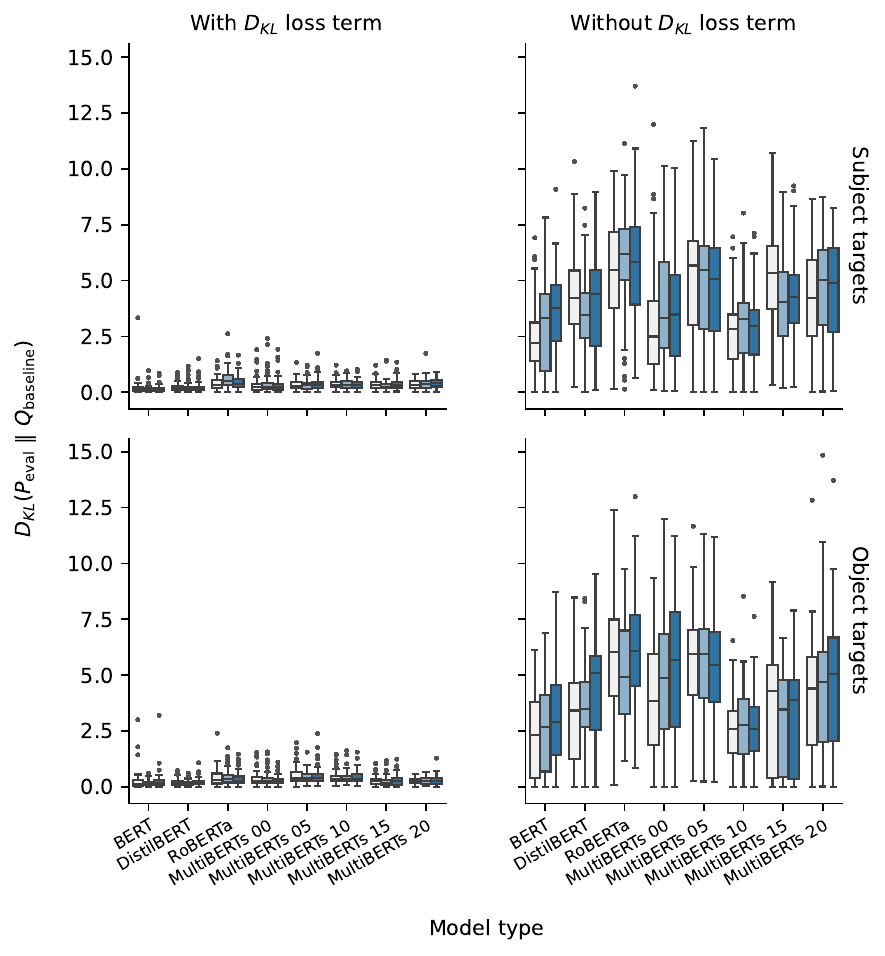}
                \caption{\label{fig:exp2-kldivs-targets} $D_{KL}$ distributions for each fine-tuned model type compared to its pre-trained version in experiment 2, calculated on the basis of 120 examples (10 per argument noun) randomly chosen from English Wikipedia that contained the nouns used as argument nouns for \textsc{blorked} in experiment 2. Each model has three associated distributions, one for each of the random seeds used in experiment 2.}\vspace{-4mm}
            \end{figure}

            Figures \ref{fig:exp2-kldivs} and \ref{fig:exp2-kldivs-targets} show  the effect of using our modified loss term, which greatly reduced $D_{KL}$ compared to using only $L_{CE}$ (the default), for randomly chosen positions of randomly chosen sentences (fig. \ref{fig:exp2-kldivs}) 
            and for positions of the subject and object argument nouns of \textsc{blorked} in randomly selected corpus sentences containing these words (fig. \ref{fig:exp2-kldivs-targets}). The range of mean KL divergences (for fig. \ref{fig:exp2-kldivs}) was $[0.578, 1.12]$ for models with the KL divergence loss term, and $[5.29, 14.6]$ without. (In both cases, this range does not include the mean KL divergence for RoBERTa, which was an outlier with means $5.53$ and $38.3$, respectively.) For comparison, we calculated the $D_{KL}$ of the BERT base uncased and DistilBERT base uncased models using the same methodology on the basis of our full 10,000 example mini-dataset, which yielded a value of $0.9$.
            
            This value constitutes a strong baseline as DistilBERT's pre-training objective is to approximate BERT's predicted probability distributions \citep{Sahnetal:2019}. The fact that post-fine-tuning $D_{KL}$ is comparable to this value suggests fine-tuning has not substantially disrupted the LLMs' linguistic knowledge.\footnote{An anonymous reviewer suggests applying a similar penalty term for each layer, following ideas from the knowledge distillation literature that \citet{gou2021knowledge} call feature\hyp based knowledge transfer.  We attempted such an addition, but it did not warrant extended investigation: it impeded the network's success on the training data and did not improve generalization to the test data.}
        
        \subsection{Results}     
            We evaluated model performance using accuracy as described in experiment 1. However, rather than comparing the probability of different nouns in the same position as in experiment 1, we instead compared the probability assigned to each of the twelve selected nouns across positions in a sentence: if noun $n$ is assigned higher probability in the appropriate position $P$ than in the other position $P'$, it is counted as correct. This eliminates the potential confound of high-frequency nouns being assigned higher probability overall. Comparing a single token's probability in different contexts reveals the extent to which the model has learned to distinguish \textsc{blorked}'s different argument positions. We also computed a confidence score as a more fine-grained assessment of model performance.
            \[
                \text{Confidence}(n) = \log p(n|P) - \log p(n|P')
            \]
            For presentational purposes, we use ``subject'' and ``object'' to refer to the underlying roles of the arguments. Though this introduces potential confusions, as the surface subject of a passive is labeled as its object, it has the advantage of collapsing across positions in which identical predictions should be made for possible nouns.   
            
            An important concern here regards the fairness of this task (raised by an anonymous reviewer). Just because a novel verb occurs with a particular noun in object position, this does not necessarily license the inference that the noun is less likely to occur in subject position, or that it may only occur in object position. Many verbs allow the same\slash similar arguments to occur in either subject or object position (e.g., \emph{meet}, \emph{like}, etc.). A lack of a learned preference might indicate not that the LLMs have failed to generalize, but rather that they have instead displayed a reasonable conservatism, not generalizing from appearance in one context to non-appearance in another. It would thus be unfair to penalize the models for having failed to generalize if they displayed no preference. But in fact, almost all of our models display clear preferences for the appearance of our target nouns in both simple actives and simple passives. Thus, we can use these preferences to evaluate whether the models are generalizing correctly.
            
            \begin{table}
                \centering\tiny
                \begin{tabular}{cr|rr|rr}
                    \toprule
                                        & \multicolumn{1}{c|}{Fine-tune} & \multicolumn{2}{c}{Active} & \multicolumn{2}{|c}{Passive}\\
                                        & \multicolumn{1}{c|}{acc.} & \multicolumn{1}{c}{SO} & \multicolumn{1}{c}{OS} & \multicolumn{1}{|c}{SO} & \multicolumn{1}{c}{OS}\\
                    \midrule
                        RoBERTa         & \tp{79.2} & \tp{66.7} & \tp{53.6} & \tp{44.8} & \tp{39.5}\\
                        BERT            & \tp{86.1} & \tp{75.5} & \tp{54.2} & \tp{54.4} & \tp{55.7}\\
                        DistilBERT      & \tp{88.9} & \tp{74.8} & \tp{49.4} & \tp{51.9} & \tp{39.9}\\
                    \midrule
                        MultiBERT 00    & \tp{80.6} & \tp{63.9} & \tp{50.8} & \tp{48.7} & \tp{40.8}\\
                        MultiBERT 05    & \tp{84.7} & \tp{80.4} & \tp{49.8} & \tp{51.0} & \tp{21.4}\\
                        MultiBERT 10    & \tp{82.6} & \tp{70.1} & \tp{55.2} & \tp{44.1} & \tp{42.7}\\
                        MultiBERT 15    & \tp{76.4} & \tp{68.8} & \tp{49.7} & \tp{53.3} & \tp{43.6}\\
                        MultiBERT 20    & \tp{79.2} & \tp{66.4} & \tp{65.2} & \tp{41.0} & \tp{43.0}\\
                    \bottomrule                    
                \end{tabular}
                \caption{\label{tab:linearorderacc} Mean conditional accuracy by model, voice, and relative linear order of argument types. Non-conditional accuracy on the fine-tuning data structure is included in the leftmost column. ``Subject'' and ``object'' arguments are defined relative to canonical active structures. SO = subject arguments linearly precede object arguments (as in simple actives); OS = object arguments linearly precede subject arguments (as in simple passives). Fine-tuning data for experiment 2 uses SO order with actives.}\vspace{-4mm}
            \end{table}
            
            The left column of Table \ref{tab:linearorderacc} reports accuracy on the fine-tuning data. Though the models do not succeed on the fine-tuning data to the same degree as in experiment 1, they perform considerably better than chance, indicating that they have learned an asymmetrical generalization for sentences like the fine-tuning data. Other columns report generalization accuracy by voice (active vs.\ passive) and linear order of \textsc{blork}'s arguments: subject preceding object (SO) or object preceding subject (OS).\footnote{For reasons of space, we do not discuss how additional structural factors affected generalization in experiment 2. In fact, we found that the patterns seen in active\slash passive generalization were similar to the patterns seen with generalization across other kinds of structural contexts. All of our data is publicly available via our GitHub repo (\url{https://github.com/clay-lab/structural-alternations}).}  Simple active sentences (most like the fine-tuning data) are SO, while simple passives are OS. Voice and argument order are decoupled in sentences like \emph{Which [obj] has the [subj] \textsc{blorked}?} (active OS) and \emph{Which [subj] was the [obj] \textsc{blorked} by?} (passive SO). While the models generalize fairly well to active SO sentences, performance is lower for other sentence types. In almost all cases, the choice of a position for a given noun is at or below chance. The exception is BERT, where generalization to other sentence types is slightly above chance, though not nearly at the level of the fine-tuning or active SO cases.
            
            We found little difference across the MultiBERT models, which failed to consistently generalize in all but active SO sentences. Indeed, in passive sentences, most models displayed below chance performance, most consistent not with a lack of preference, but rather with a preference to place arguments in the opposite of their correct positions. While chance performance on its own could have been due to problems with our task, the preferences in simple actives and simple passives indicate that the models have learned preferences, but not always the ones that accord with the correct mapping between active and passive structures.
            
            Generalization seems to be guided by the relative linear order of the arguments and the verb: if the subject argument occurs before the verb, which in turn precedes the object, as in the fine-tuning data, generalization is quite good. In both active and passive sentences, conditional accuracy always decreases for sentences with OS argument order, as well as passive SO sentences, where both subject and object precede the passive verb. We hypothesize that this pattern of performance results from the models' using a linear heuristic to generalize across sentence types.
            
\begin{figure}[t]
                \centering
                \includegraphics[width=0.75\linewidth]{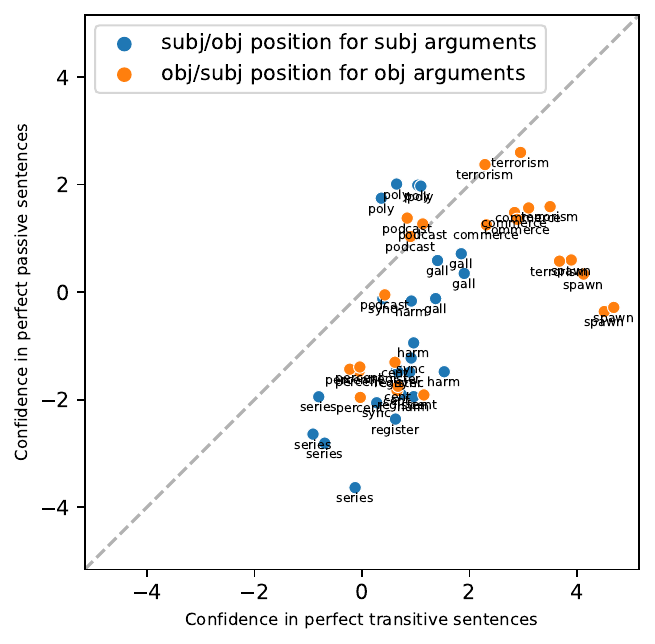}
                \caption{\label{fig:exp2single} A single BERT's performance on perfect active transitive (e.g., \emph{The [subj] has always \textsc{blorked} the [obj].}) and perfect passive (e.g., \emph{The [obj] has always been \textsc{blorked} by the [subj].}) sentences. Fine-tuning data for experiment 2 consisted of perfect transitive sentences. The diagonal line indicates equal confidence (i.e., perfect generalization) across sentence types.}\vspace{-4mm}
            \end{figure}

            We investigated this behavior at a more fine-grained, individual structure level as well. Figure \ref{fig:exp2single} compares confidence for one representative BERT model between perfect transitive sentences (i.e., sentences that match the fine-tuning structure) and corresponding perfect passive sentences. Although the model has learned the correct positions of the arguments in perfect transitive sentences (almost all points are positive on the $x$-axis), it fails to generalize this to corresponding passive sentences (points are not systematically positive on the $y$-axis), though generalization performance on objects is noticeably better than on subjects (conditional subject acc.: $35.0\%$; conditional object acc.: $65.0\%$; overall: $50.0\%$).
            
            \begin{figure}[t]
                \centering
                \includegraphics[width=0.75\linewidth]{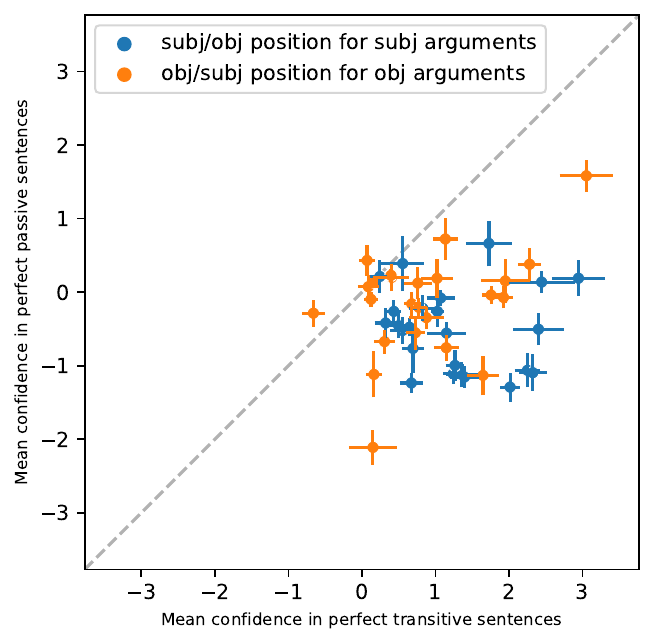}
                \caption{\label{fig:exp2multi} Mean performance from 24 models (BERT; DistilBERT; RoBERTa; MultiBERT 00, 05, 10, 15, 20 $\times$ 3 distinct model-specific argument sets) on perfect transitive and perfect passive sentences. Each point is the mean confidence score for an argument type taken across individual argument tokens and sentences of that type for a given model run. Bars are standard errors.}\vspace{-4mm}
            \end{figure}
            
            Figure \ref{fig:exp2multi} zooms out a bit, with each point encoding the confidence scores of a single pre-trained model with one of three random seeds. The performance of each pre-trained model is represented by two points, one for subjects and one for objects, averaged across distinct nouns for each argument type and a set of perfect and passive sentences. We see, once again, no tendency to generalize systematically from active to passive.
            
\begin{figure}[t]
                \centering
                \includegraphics[width=\linewidth]{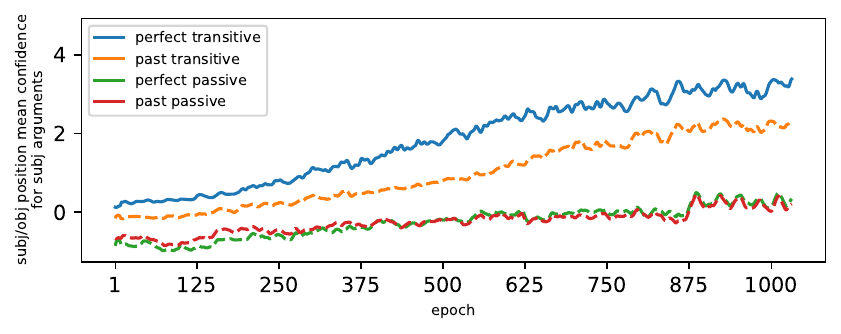}\vspace{-1.5em}
                \includegraphics[width=\linewidth]{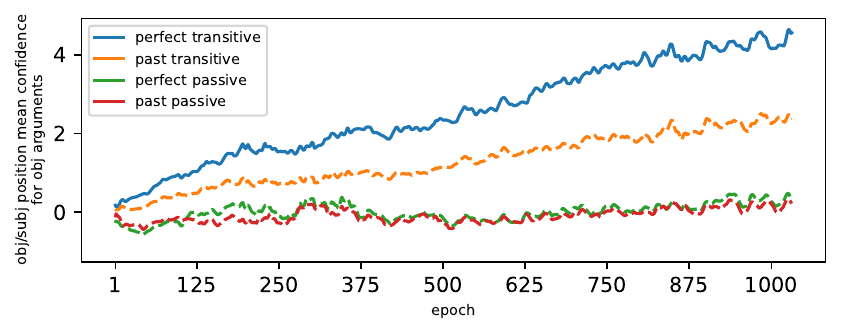}
                \caption{\label{fig:exp2metrics} Mean confidence during fine-tuning for transitive (active) and passive sentences for each argument type. Perfect transitive is the fine-tuning data with dropout disabled; others are validation set subparts. While performance on actives increases substantially, performance on passives increases only a little over the entire course of fine-tuning.}\vspace{-4mm}
            \end{figure}

            We might also detect Type 2 generalizations through the course of learning: changes in the confidence of argument predictions should run in parallel across structures that are linked by a Type 2 generalization. To test for this, we tracked mean confidence scores for different sentence and argument types over the course of fine\hyp tuning for each weight update. Figure \ref{fig:exp2metrics} gives a representative plot for a BERT model tuned until convergence. Perfect and past tense actives move together. In contrast, perfect and past passives show little improvement for objects, and only a tiny improvement for subjects that stabilizes below actives (around 0, indicating chance performance).
            
        \subsection{Investigating over- and undergeneralization}
            Experiment 2's results show that appear to overgeneralize, in that they generalize linearly to from active transitives to passives; and undergeneralize, in that they fail to generalize from active SO sentences to active OS and passive SO sentences.
            
            There is a caveat regarding the failure of overgeneralization: not all verbs that appear in SVO active sentences can passivize.\footnote{We thank an anonymous reviewer for raising this point.}
            \begin{exe}
                \ex{\begin{xlist}
                    \ex[*]{A car is had by a driver.}\label{unpasshave}
                    \ex[*]{Five dollars are cost by five tickets.}
                    \ex[*]{The boy is resembled by the girl.}
                \end{xlist}}\label{exp2pass}
            \end{exe}
            While there are relatively fewer verbs like these than verbs that do passivize, they represent an important confound. This is because, in the absence of explicit positive evidence that \textsc{blorked} can passivize, the models may have instead come to treat it as belonging to the class of unpassivizable verbs like \emph{have}, \emph{cost}, \emph{resemble}, etc. If this were the case, we would be testing the models on apparently ungrammatical sentences, where it is not clear what the ``correct'' behavior should be.
            
\begin{figure*}[t]
                \centering
                \includegraphics[width=\linewidth]{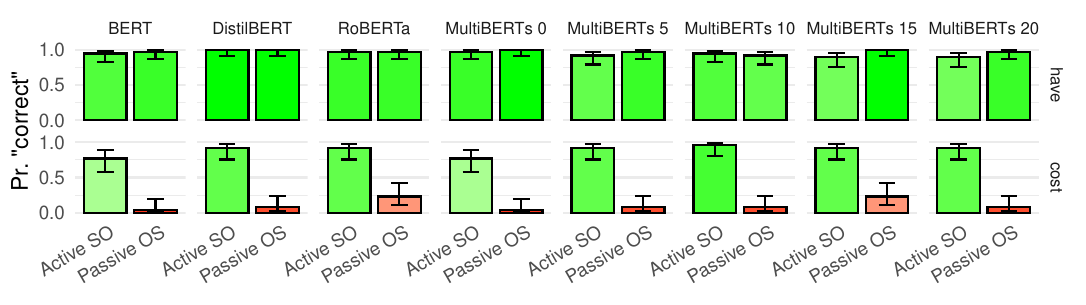}
                \caption{\label{fig:exp2overgen} ``Accuracy'' across argument tokens for active and passive sentences with unpassivizable verbs. We define ``accuracy'' as making the predictions expected if the verbs could passivize. Bars are 95\% CIs.}\vspace{-4mm}
            \end{figure*}

            To investigate this, we extracted predictions from the baseline versions of the models used in experiment 2 for the argument positions of simple passive sentences like those in (\ref{exp2pass}) and their (grammatical) simple active counterparts for a variety of good subject and good object nouns. We determined good subject and object nouns using the \href{https://www.english-corpora.org/iweb/}{iWeb corpus} by searching for partial strings matching the active versions of our sentences (e.g., \texttt{"}\emph{A NOUN has a}\texttt{"} and \texttt{"}\emph{has a NOUN .}\texttt{"}). We chose from among the $2000$ most frequent nouns for one argument that did not appear among the $2000$ most frequent nouns for the other argument that were tokenized as whole words. We then extracted predictions for each token for each position from the models for active and passive sentences for \emph{have} and \emph{cost}, to see what the models' predictions were for sentences with unpassivizable verbs. If these predictions differ from the predictions for the arguments of \textsc{blorked}, we could tentatively conclude that the models did not treat \textsc{blorked} as unpassivizable. We report ``accuracy'' in fig. \ref{fig:exp2overgen}, where being correct is defined as though the verbs were passivizable; a prediction was accurate if the token was more probable in the expected position than in the alternative position.
            
            We notice a clear difference between an unpassivizable verb like \emph{have}, which the models seem to treat as passivizable; and an unpassivizable verb like \emph{cost}, for which the models generalize in the linear fashion we saw with \textsc{blorked}. This unfortunately leaves us with an unclear result: perhaps the models are treating \textsc{blorked} as belonging to the subset of unpassivizable verbs like \emph{cost}, rather than to the subset of those like \emph{have}. We note that a possible reason that the models may treat unpassivizable uses of \emph{have} as passivizable is that there are some passive uses of \emph{have}, such as \emph{A good time was had by all}, despite this being restricted to particular contexts (compare (\ref{unpasshave})).
            
            We thus cannot decisively say that the models have failed in their treatment of \textsc{blorked} in passives: one \emph{could} consider this a case of learning an incorrect linear generalization about the distribution of \textsc{blorked}'s arguments; but on the other hand, it is also possible that the model has instead learned that \textsc{blorked} is an unpassivizable verb like \emph{cost}, and thus the apparent failure is actually conservatism in the face of ambiguous evidence.
            
            Despite this complication, we still do not consider this to compromise our overall view that these LLMs in general fail to represent Type 2 knowledge. In particular, we did not only consider passives in experiment 2, but also, crucially, active OS sentences involving \abar{A}-movement like clefting, \emph{wh}-movement, etc. As far as we know, no verb disallows such movement of its arguments, even unpassivizable verbs: e.g., \emph{It is a car that a driver has}, \emph{It was five dollars that five tickets cost}). But in table \ref{tab:linearorderacc} we saw that the models typically failed to generalize to active OS sentences, with performance hovering around chance. Even if we do not want to consider the results for passive OS sentences to represent overgeneralization in light of fig. \ref{fig:exp2overgen}, we consider the results for active OS sentences to represent undergeneralization, in that they display no knowledge of Type 2 generalizations that relate active SO and active OS sentences.
        
        \subsection{Discussion}
            Experiment 2's results suggest the LLMs under study do not encode Type 2 generalizations that cut across active and passive sentences for a novel verb seen only in active sentences. However, these LLMs do show signs of generalizing across structures, from the fine-tuning structure to other active SO sentences (Table \ref{tab:linearorderacc}). This requires a Type 2 generalization that links, e.g., perfect and past-tense sentences, positive and negative sentences, or declaratives and polar questions. However, this generalization need not be structure-sensitive, unlike Type 2 generalizations found in natural language. Rather, LLMs' favored generalizations appear to collapse contexts based on a linear template---in this case, N$_1$ $\prec$ V $\prec$ N$_2$. In passive OS sentences, this leads to the wrong prediction: the initial noun is predicted to have the same distributional constraints as active subjects, while the \emph{by}-phrase noun, interpreted as N$_2$, has the  distributional constraints of active objects, yielding below chance performance in almost all cases. Active OS and passive SO sentences, with N$_1$ $\prec$  N$_2$  $\prec$ V order do not match the template, so fine-tuning provides no useful guidance, leading to chance performance.
            
            The preference for linear generalizations recalls previous findings about Transformers' inductive bias \citep{petty2021transformers}, but conflicts with more recent results with large pre-trained sequence-to-sequence models \citep{mueller-etal-2022-coloring}.  Carrying out Type 2 generalizations goes beyond the inductive bias necessary for structure-sensitive mappings, as it requires drawing inferences independent of lexical content.
    
    \section{Related work}
        Past work has probed token embeddings for knowledge of argument structure \citep{Kannetal:2019,Warstadtetal:2020,Tenneyetal:2019b,Tenneyetal:2019a,Pavlick:2022,ZhudeMelo:2020,SasanoKorhonen:2020}. Other work has focused on neural networks' ability to predict the likelihood of a verb or noun in forms of an argument structure alternation \citep{ChowdhuryZamparelli:2019,WarstadtBowman:2019,Hawkinsetal:2020,YoshidaOseki:2022,Loaicigaetal:2021,PettyWilsonFrank:2022a,Methenitietal:2020}, and whether LLMs distinguish plausible from implausible argument-role mappings in role-reversal sentences \citep{Ettinger:2019}. Though revealing of Type 0 knowledge, that work does not address whether LLMs can apply such knowledge productively, which is what drives our study. Experiment 1 adapts \citet{PettyWilsonFrank:2022a}'s approach to probe Type 1 knowledge, but we consider different verbs in a wider range of structures, allowing us to assess this knowledge in a more refined way along dimensions such as linear order and movement type. However, our work extends these previous investigations by considering in more detail than this work how the models achieved success by examining the learned embeddings of the novel words, finding that the possibility of assimilation seemed to underlie success, and by examining Type 2 knowledge directly.
	    
        \citet{Thrushetal:2020} found BERT predicted unseen in-group pairings of novel verbs with novel nouns to be more likely than unseen out-group pairings, showing BERT can generalize selectional preferences across verbs. They also showed BERT can generalize the use of novel verbs that occurred in contexts compatible with one form of an argument structure alternation to the other context (e.g., a novel verb V in a DO context to V in a PD context). While this work also focuses on the generalization of selectional preferences, we examine structures beyond the argument structure alternation itself, giving us a better assessment of BERT's abilities with Type 2 generalization over structurally defined relations between contexts.
	    
        Methodologically, \citet{Lasrietal:2022}'s work is most similar to ours. They reinvestigate \citet{Goldberg19}'s claim that BERT can generalize its knowledge of subject-verb number agreement in English to novel structures and lexical items. \citet{Lasrietal:2022} show that Goldberg's results hold only for simple sentences and those whose lexical patterns match naturally occurring data. BERT fails to generalize correctly to grammatical but nonsensical sentences in more complex structures, where as little as a single attractor noun intervenes between the correct agreement target and the verb. \citet{newman-etal-2021-refining} came to similar conclusions on the basis of a different approach to subject\hyp verb agreement, finding that models were less likely to predict correct agreement morphology for verbs that were improbable in particular contexts. This work provides a distinct but complementary perspective regarding our interpretation of our results, given our observation that models seem to do well only when they can assimilate novel words to existing words (experiment 1, but not experiment 2). This suggests that a general reliance on surface\hyp level similarities to the pre\hyp training data underlies success in generalization. \citet{Lasrietal:2022} show that this is true of the morphological fact of subject\hyp verb agreement, and we show it is true for the structural fact of argument structure alternations (and movement more generally). 
    
    \section{Conclusion}
        We investigated the performance of transformer\hyp based language models on two tasks requiring structural generalization.
        
        Experiment 1 revealed that LLMs were generally able to succeed at predicting novel arguments to known verbs in correct positions in structures different from those in the fine\hyp tuning data. An investigation of the learned embeddings of the novel tokens showed that the LLMs can use a syntactic bootstrapping strategy \citep{Gleitman:1990, Gleitmanetal:2005}, where information about syntactic distribution is used to inform hypotheses about meaning. Specifically, LLMs assimilate the embeddings of novel tokens to those of existing words that share their distribution. If an LLM constrains distributions on the basis of embedding subspaces, the distribution of the novel token in a previously observed context in which it has not appeared can thereby be driven by its similarity to other tokens, giving rise to successful Type 1 generalization.
        
        Experiment 2 found that, when this strategy was unavailable, models were able to generalize successfully only when the relative linear order of the novel verb's arguments matched in the fine\hyp tuning and test data, but degraded noticeably for reversed orders. We conclude that the models we investigated can represent Type 2 knowledge across structural contexts, but that the kind of Type 2 knowledge they represent is based on surface level properties like the relative linear order of corresponding elements.
        
        While this study focuses on one microscopic aspect of linguistic knowledge, the lessons it offers about what LLMs are capable of are significant. Impressive as their performance on a wide range of downstream tasks is they appear to have limited ability to generalize in a human\hyp like structure\hyp sensitive fashion. Appearances to the contrary may stem from the separate learning of distributional properties across multiple contexts. The necessity for such redundant learning imposes a high demand on the amount of training data that is required for LLMs, which we expect can be satisfied for only a very few high resource languages.  

\section*{Acknowledgments}

For helpful comments, critiques and suggestions, we would like to thank Tal Linzen, Alec Marantz, the members of the Computational Linguistics at Yale (CLAY) lab and the Computation and Psycholinguistics lab at NYU, and the audiences at the Jabberwocky Words in Linguistics workshop and at the University of Toronto, Northwestern University, Stony Brook University, UCLA and Heinrich Heine University D\"usseldorf. The comments and suggestions of the anonymous TACL reviewers were also invaluable. This work was made possible by support from National Science Foundation grant BCS-1919321.
    
	\bibliography{tacl2021}
	\bibliographystyle{acl_natbib}
\end{document}